%% file: acl2023.tex
\pdfoutput=1
\documentclass[11pt]{article}

\usepackage[]{ACL2023}

\usepackage{times}
\usepackage{latexsym}
\usepackage{enumitem}

\usepackage[T1]{fontenc}
\usepackage[utf8]{inputenc}

\usepackage{microtype}
\usepackage{booktabs}
\usepackage{graphicx}
\usepackage{multirow}

\usepackage{inconsolata}
\usepackage{array,multirow,graphicx,float}

\title{Summarization from Leaderboards to Practice: Choosing A Representation Backbone and Ensuring Robustness}

\author{David Demeter$^{1}$ \quad 
        Oshin Agarwal $^{2}$\thanks{~~Work done at Adobe Research} \quad
        Simon Ben Igeri$^{1}$ \quad 
        Marko Sterbentz$^{1}$ \quad \\
        \bf Neil Molino$^{3}$ \quad 
        John M. Conroy$^{3}$ \quad
        Ani Nenkova$^{4}$ \\\\
        $^{1}$Northwestern University \quad
        $^{2}$University of Pennsylvania \\
        $^{3}$IDA/CCS \quad
        $^{4}$Adobe Research \\
        {\tt \{ddemeter,simon.benigeri,markosterbentz2023\}@u.northwestern.edu} \\
        {\tt oagarwal@seas.upenn.edu} \quad
        {\tt \{npmolin,conroy\}@super.org} \quad {\tt nenkova@adobe.com}
}

\begin{document}
\maketitle
\begin{abstract}
Academic literature does not give much guidance on how to build the best possible customer-facing summarization system from existing research components. Here we present analyses to inform the selection of a system backbone from popular models; we find that in both automatic and human evaluation, BART performs better than PEGASUS and T5. 
We also find that when applied cross-domain, summarizers exhibit considerably worse performance. At the same time, a system fine-tuned on heterogeneous domains performs well on all domains and will be most suitable for a broad-domain summarizer. Our work highlights the need for heterogeneous domain summarization benchmarks. We find considerable variation in system output that can be captured only with human evaluation and are thus unlikely to be reflected in standard leaderboards with only automatic evaluation. 
\end{abstract}

\section{Introduction}

Academic papers on automatic document summarization have been published since the 1950s \cite{luhn1958automatic} but broadly applicable summarizers not constrained by document type have only recently become widely available.\footnote{\url{https://ai.googleblog.com/2022/03/auto-generated-summaries-in-google-docs.html}, \url{https://quillbot.com/summarize}, \url{https://smmry.com}} The literature contains a wealth of information on model architectures for summarization, yet it remains hard to decide from published evaluations which are ``the best'' components (data and model) for a good quality customer-facing summarizer.

Here we make the idealized assumption that size and inference cost of the models are not an issue. We seek to find the best backbone for a neural summarizer from freely available research components, producing the best summaries, and a confirmation that the summarizer will work well for varied types of input documents. 
For this purpose, we fine-tune and evaluate popular off-the-shelf pre-trained models BART \cite {Lewis2020BARTDS}, PEGASUS \cite{Zhang2020PEGASUSPW} and T5 \cite{JMLR:v21:20-074} on six summarization datasets.
We also create mixed training datasets with a balanced representation of each of the domains. 
We find that fine-tuning on mixed-domain text, smaller in size than most of the in-domain training set, yields a robust system performing on par with models fine-tuned on the order of magnitude more data when tested in-domain.

In addition to evaluation with automatic metrics, we conduct a human evaluation. BART summaries were preferred more often than those produced by PEGASUS and T5.
Additionally, summaries generated with BART trained on mixed data are preferred over those generated with BART trained on the most popular summarization research dataset, CNN/Daily Mail, even though the mixed-domain dataset is the smaller of the two. 
Summaries from this system were even preferred over those produced by BART, fine-tuned on in-domain data matching each test sample. This preference is not captured by automatic metrics. BART fine-tuned on the mixed domain, and often produced summaries deemed more informative than the human reference for the respective input. This was not the case for summarizers obtained by fine-tuning using data from a single source.

\input{dataset_statistics.tex}

\section{Related Work}

Some hints that domain robustness is a problem but that summarizers can to an extent generalize across domains are found in the literature. \citet{yu-etal-2021-adaptsum} observe catastrophic forgetting during domain adaptation via continual pre-training. This is concerning if the goal is to have a robust system that serves multiple domains. They do not explicitly measure how much systems degrade when evaluated out of domain, though it is implied by the task and results that there is degradation. 

There are a few direct studies of summarization cross-domain robustness. \citet{sandu-etal-2010-domain} tested if meetings summarization data is useful for email summarization. They find that training on email data is best, but in the absence of such data training on meetings is helpful. \citet{bar-haim-etal-2020-quantitative} train a system for extracting key points on argumentation datasets and then evaluate the same system on municipal surveys and user reviews. The systems perform well, exhibiting robustness. In our work, we carry out a similar evaluation but we examine the robustness of abstractive summarizers on a diverse set of datasets.

These findings on cross-domain robustness are encouraging and in line with \citet{hua-wang-2017-pilot}'s findings that some of the capabilities for identifying summary-worthy content are transferable between domains.  They study news and opinion piece summarization for texts drawn and find that a model trained on out-of-domain data can learn to detect summary-worthy content, but may not match the generation style in the target domain. 
Stylistic markers of a domain i.e. as in typical phrasing used to talk about certain topics are not captured.

\section{Experimental Design}

Abstractive summarizers generate a short plain text summary capturing the main points of a longer text. The current state-of-the-art models for the task are transformer-based encoder-decoder text-to-text models, such as BART \cite {Lewis2020BARTDS}, PEGASUS \cite{Zhang2020PEGASUSPW} and T5 \cite{JMLR:v21:20-074}. The models are pre-trained on large general-purpose corpora followed by fine-tuning on specific summarization datasets. 

\subsection{Pre-trained Models}

We work with pre-trained BART, PEGASUS, and T5 models, using the model and implementation in Huggingface \cite{Wolf2020TransformersSN}. We then fine-tune these for summarization ourselves, on six summarization datasets.  All three models use a sequence length of 512 tokens and truncate inputs longer than this. Further details for each model can be found in the appendix.

\subsection{Datasets}

We use six datasets covering diverse domains, namely arXiv \cite{Cohan2018ADA}, billsum \cite{kornilova-eidelman-2019-billsum}, CNN/DailyMail \cite{Hermann2015TeachingMT}, GovReport \cite{Huang2021EfficientAF}, Pubmed \cite{Cohan2018ADA} and Reddit TIFU \cite{kim-etal-2019-abstractive}.  
The texts in each dataset differ by length and stylistic features such as formality of style, letter casing, and punctuation. These distinctions are compelling for exploring cross-domain robustness. Statistics on domain, length, and summary source are shown in Table \ref{tab:datasets}. We use the dedicated training set to fine-tune the three models we compare and a balanced subset of 250 samples from each domain for evaluation.\footnote{Inference time is approximately one week to generate summaries for the full test sets on a machine configured with three Quadro-RTX 8000 GPUs.}

We construct one additional training dataset derived from mixing the original sources (\emph{Mixed}). We uniformly sample each of the six publicly available datasets up to the number of individual examples in the dataset with the fewest observations (GovReport). This results in a training set with 105k observations. The mixed-domain dataset is larger than BillSum, GovReports and Reddit, but smaller than the training split of the other three datasets. We fine-tune models on the mixed domain dataset to evaluate if robustness can be improved with a data-only solution, where the system is exposed to heterogeneous fine-tuning data. We use the mixed domain test set as a single test set for evaluating summarizer robustness. 

\subsection{Evaluation Settings}

We explore three fine-tuning and testing configurations.  {\em In-domain} testing is when the source of the test sample matches the fine-tuning source, as is conventionally done in summarization research. 
{\em Cross-domain} testing is when a summarizer fine-tuned on one source of data is used to generate summaries for another source.
We also perform {\em mixed-domain} testing, in which we evaluate the summarizers fine-tuned on mixed-domain data on each of the six summarization datasets. 

{\em In-domain} summaries align well with prior published results based on standard datasets, developed for convenience and fast evaluation. {\em Mixed-domain} evaluation and summarizers are the most relevant to real-world use cases among the regimes studied in this work. 

\input{avg_auto_metrics.tex}

\input{detail_auto_metrics.tex}

\section{Automatic Evaluation}

We first evaluate the summarizers using three automatic metrics: ROUGE-2 \cite{Lin2004ROUGEAP}, sacreBLEU \cite{Post2018} and BERTscore \cite{bert-score}.
The goal of this evaluation is to glean insights about system performance to inform the choice of specific comparisons that can be done with human evaluation. 

We show the average in-domain and the average cross-domain scores for each model in Table~\ref{tab:averages}. 
Based on the automatic scores, BART is the best backbone model, with the best performance on all three automatic evaluations both in in-domain and in cross-domain evaluation. PEGASUS is better than T5 in in-domain evaluation, but both are similar in cross-domain evaluation. 
All three automatic scores are much lower for cross-domain evaluation compared to in-domain evaluation, suggesting that domain robustness poses a problem for a practical system. The drop in ROUGE2 and BLEU is much higher than that in BERTscore.

We also show the average automatic scores on the six test datasets with BART trained on different settings (Table \ref{tab:averages_data}). The in-domain score reports the average of the six models trained on each of the datasets and evaluated in-domain. CNN represents a single model trained on just CNN and evaluated on each of the six datasets. Similarly, mixed-domain is a single model trained on the mix-domain training set and evaluated on each of the test sets. All three scores show that in-domain is better than mixed-domain, which in turn is better than CNN. CNN is the largest dataset so the scores are not dependent on the training data size, rather it is the domain that matters.

For a detailed view, in Table~\ref{tab:cross_domain_agg}, we show the in-domain scores along with the respective average deterioration in cross-domain evaluation. The cross-domain panel lists for the training set, the average of the difference between the score on the in-domain test data and that on each of the cross-domain test datasets. The smaller this difference is, the more robust the summarizer is in cross-domain evaluation. The summarizer fine-tuned on mixed-domain data has the smallest cross-domain degradation on all three automatic evaluation scores, for all pre-trained models. Training on mixed-domain data yields the most robust summarizer.

\section{Human Evaluation}
Automatic evaluations consistently indicated that \emph{(i)}~BART produces better summaries than T5 and PEGASUS across the six domains we study, and \emph{(ii)}~the summarizer trained on mixed domain data is the most robust to domain changes. To confirm this finding, we also conduct a manual human evaluation. We sample 10 examples from each domain, for a total of 60 documents\footnote{Our initial plan was to run a human evaluation on larger samples test sets but based on our initial exploration, we no longer believe this is a meaningful endeavor. We discuss this in \S \ref{sec:experience}}. Each example has a human reference summary and 5 automatic summaries. The same trends for automatic scores are observed for these 60 documents as the 1500 documents in the last section.

\subsection{Evaluation Setup}
Three of the authors carried out two rounds of evaluation. In the first round, we compared the human summaries to summaries produced by BART, T5 and PEGASUS fine-tuned on the mixed-domain training set. 
The goal of this comparison is to find which of the models produced the best summaries. Overall, BART was the most preferred system, consistent with automatic evaluation.

In the second round, we compared three BART summarizers: fine-tuned on the mixed domain; fine-tuned on CNN/Daily Mail; fine-tuned on data matching the input source. 
Given the automatic evaluation, we expect that the in-domain summarizer will be best. However, the mixed-domain BART summarizer was the most preferred one.

The judges were first asked to read all four summaries for a given input, without seeing the input itself. The human summary was always placed first in the interface and marked as human. The other three summaries were displayed next, presented in random order for different inputs and listed as Summary A, B, and C, concealing the system that produced the summary. The judges were asked to compare the relative quality of the human and the machine summaries: ``Do some automatic summaries provide better content? 5 (a lot of better content) to 1 (no better content)''. 

After the judges read all four summaries and answered the above question for the human summary, they were shown three consecutive pages, each listing one of the summaries and the following questions:

\begin{description}

\item[readability] Is the summary easy to read (formatting, length, style) 5 (very easy to read) to 1 (not at all easy to read)?

\item[recall] Does the summary provide good information 5 (a lot of good info) to 1 (no good info)?

\item[precision] Does the summary have unnecessary information 5 (lots of unnecessary info) to 1 (no unnecessary info)?

\item[hallucination]  Does the summary contain apparent hallucinations 5 (no discernable hallucinations) to 1 (obvious hallucinations)?

\item[orthography]  Is the summary formatted according to the rules of English? (yes/no)

\item[repetition]  Does the summary have repetitions? (yes/no)

\end{description}

\subsection{Comparing Model Architectures}

In the first round, BART trained on mixed domain data emerged as the clearly preferred model over T5 and PEGASUS. Table \ref{tab:round1} shows the average rater score for the mixed domain test set summaries produced by each model. For precision and repetition, a lower score is better. For all other dimensions, a higher score is better. BART has a higher score that denotes that summaries conform to the rules of English orthography when compared to other models, though the absolute score is low. BART fine-tuned on mixed-domain data is also rated as having summaries with the best information recall and readability. It does not produce summaries with repeated content within the summary, while T5 often and PEGASUS occasionally do. BART summaries have the least amount of unnecessary information i.e. high precision for information content. The manual evaluation confirms the findings from the automatic evaluation.
PEGASUS is rated as the next choice, over T5 on all dimensions. These findings align with the automatic evaluation but provide considerably more nuance with respect to the dimensions in which the summaries differ. 

\input{round1_human.tex}

Hallucinations were rarely detected for any of the systems, though the judgment was made on the basis of the human summary alone, rather than the full input text. T5 produces the most apparent hallucinations. It also produces significantly more unnecessary content than the other models and its summaries often contain repetitions. Empirical benchmarking presented in published research had not prepared us to expect these.

Orthography is problematic for all models, with less than half of the summaries rated as acceptable. In many cases, the summarizers faithfully imitate the incorrect formatting, tokenization and orthography of the fine-tuning data for each domain and the rating often reflects this aspect of system behavior\footnote{Only the CNN-Daily Mail fine-tuning dataset follows orthography conventions.}. The datasets are developed for research purposes, without forward planning to present the results in front of human readers. Most summaries also end mid-sentence, which is jarring when summaries are intended for people.

\subsection{Comparing Training Data}

\input{round2_human.tex}

Next, we repeat the same evaluation protocol to compare a BART summarizer fine-tuned on three different types of datasets. In round 2 evaluation, BART fine-tuned on mixed data was rated best for the information its summaries contained and as having the least unnecessary content. 

In this second round of evaluation, {\em the human ratings revealed preferences different from what the automatic scores suggested}. The expectation from the automatic evaluation was that the in-domain system would produce the best summaries, possibly with a difference that is not statistically significant. This expectation does not bear out in the human evaluation. The mixed-domain BART system has higher readability scores than the in-domain system, has better information recall as well as precision, and produces more reasonable orthography. BART fine-tuned on mixed-domain is better than the in-domain system---a strong result with practical significance.

BART fine-tuned on CNN-DM produces the most readable summaries also following English orthographic rules, but these summaries contain the least useful information, with a point and a half drop on the five-point scale compared to the mixed-domain system. It also generates much more unnecessary information, with a difference of one whole point on the five-point scale. Ideally both the summary content will be good and the text will be readable. In our evaluation, we find that the system that produces the most readable summaries generates poor summaries content-wise. If forced to choose one, the system fine-tuned on mixed-domain will be the uncontroversial choice.

\subsection{Automatic Summaries Better than Human Reference}

The superiority of the summarizer fine-tuned on mixed-domain data also emerges in comparison with the human reference summary. The mixed-domain system produced a summary rated higher than the human summary for 18 of the 60 examples, while the in-domain system did so for only 5.

The BART-large model fine-tuned on mixed-domain was the most preferred summarizer in our manual evaluation. We found that it often produced summaries judged to be better than the human reference summary for the same document. Table~\ref{tab:better} shows the number of documents, out of 10, where the automatic summary was given a higher score than the respective human summary by at least two judges. The model fine-tuned on the mixed-domain data had the overwhelming share of summaries which provided better content than the human summaries. While such summaries were present in each of the six domains, CNN/Daily Mail was the domain with the largest, followed by Reddit. We give samples of such summaries in the appendix of the paper. This summarizer is not only better than other alternatives we studied, but it is also at times better than human summaries in domains where the human summary is just a teaser to invite a full reading of the text.

\begin{table}[t]
\small
\centering
\setlength{\tabcolsep}{4pt}
\begin{tabular}{l*3r}
\toprule
&  in-domain &  CNN-DM &  Mixed \\
\midrule
arXiv   & 1 & 0 & 2 \\
BillSum & 0 & 0 & 2 \\
CNN     & 0 & 0 & 8 \\
Gov     & 1 & 0 & 1 \\
PubMed  & 1 & 0 & 1 \\
TIFU    & 2 & 3 & 4 \\ \midrule
All     & 5 & 3 & 18 \\ \bottomrule
\end{tabular}
\caption{Number of test examples for which a BART summary was given an information recall score greater than that for the human summary by at least two annotators, indexed by domain and model.}
\label{tab:better}
\end{table}

\subsection{Human Summary Evaluation}
\label{sec:experience}

The manual evaluation was a difficult and frustrating experience. To give a sense of the problem, we show in Table \ref{tab:HumanRead} the readability scores for \emph{the human summaries} across domains, broken down by annotator. The most readable were the CNN/Daily Mail, the only cased domain, while the least readable were arXiv and PubMed, which were not only lowercased, but also contained math symbols replaced by templates. 
The government reports were excruciatingly hard to read in plain text. They are typically long, around 500 words. On the government website, these were formatted in three or more paragraphs, with some visual support in the form of a graph or chart to help in understanding. Learning to generate automatic summaries of such length without segmenting the text into paragraphs is probably a wasteful effort because people are not likely to read the plain text output. 

\begin{table}[t]
\small
\centering
\setlength{\tabcolsep}{4pt}
\begin{tabular}{lrrrrrr}
\toprule
{\bf Expert} & arXiv & BillSum & CNN & Gov & PubMed & TIFU \\
\midrule
{\bf A} & 2.5 & 3.4 & 5.0 & 2.8 & 3.0 & 5.0 \\
{\bf B} & 3.9 & 4.9 & 4.8 & 4.5 & 3.8 & 3.8 \\
{\bf C} & 3.8 & 4.5 & 5.0 & 4.0 & 3.9 & 4.0 \\
\bottomrule
\end{tabular}
\caption{Average readability scores of human summaries by each human annotator.}
\label{tab:HumanRead}
\end{table}

Annotator A gave much lower scores to the human summaries for all but the CNN and Reddit domains. In a post hoc discussion, they shared that they were reading as if the task is to tell in their own words what the text is about. The other two annotators in contrast were mostly skimming, not looking for deep comprehension. Superficial reading is unlikely to be sufficient in tasks where annotators are asked to compare the content quality in two summaries. Similarly, a person would be unable to make that judgment if they cannot understand what the text is about. The process was tedious, despite the fact that our human annotators were researchers with considerable experience in summarization. In light of these considerations, it is hard to imagine that it is ethical to crowdsource evaluations except for the news and Reddit domains. These are however the least representative of documents people may be reading for their work, where a summarizer can be helpful.

Despite the difficulty of reading the summary text, on average for the entire test set the human evaluation scores are remarkably consistent. BART fine-tuned on mixed-domain data was evaluated in Round 1, as well as in Round 2. The first columns in Tables \ref{tab:round1} and \ref{tab:Round2} are the average human ratings for the same summaries. The differences are minor, and all conclusions hold if the first columns in the two tables were swapped.

\section{Conclusions}
We study the cross-domain robustness of neural summarizers. We find that models fine-tuned on only one domain suffer cross-domain deterioration of performance.
We find that BART is the best pre-trained model for summarization. It is especially effective when fine-tuned on mixed-domain data. In the human evaluation, this summarizer is rated as producing better summaries than an in-domain summarizer and often produces summaries better than the human summary. This is not reflected in the automatic scores and will therefore not be captured by leaderboards.
We also find that most existing datasets do not support efforts toward developing a customer-facing summarizer. The data is poorly formatted and hard to read, so the resulting summaries are unlikely to lead to a delightful customer experience and are hard to read in manual evaluation. Much like the Google team that deployed the auto-summarization feature, we conclude that high-quality, and heterogeneous, fine-tuning data will be necessary to develop such a system.

\section{Limitations}

This work presents an expansive analysis of the cross-domain robustness of neural summarizers using automatic metrics and human evaluations.  The test sets for summarization datasets selected for our analysis range from about 900 to 12,000 observations, making exhaustive manual evaluation infeasible.  Instead, we elect to evaluate the first 250 observations from each dataset.  While we believe this sample is sufficient to be representative of the whole dataset, we recognize that a larger-scale human evaluation using crowd-sourced workers may be beneficial.  Our human evaluations are created with only three annotators.  

A larger-scale evaluation with a diverse set of crowd-sourced workers would also address this potential issue.  In addition, annotators only compare machine-generated summaries with human ones when performing our human evaluations and do not work with the original passage.  While comparing summaries with original passages may be ideal, some datasets' length and technical detail make this difficult, even with crowd-sourced workers.

We work with only three neural summarizers and in one size per model.  These summarizers are available in multiple sizes models, and other summarization models are available.  We elected to forgo these because we are studying cross-domain performance in general rather than trying to explain how model-specific differences manifest themselves in performance.  Lastly, we worked with only six publicly available summarization datasets and constructed the Mixu dataset using uniform sampling on each dataset.  While we could have studied a larger number of datasets, we believe that the diverse nature of our selections yields a representative analysis.

\bibliography{anthology,custom}
\bibliographystyle{acl_natbib}

\appendix

\section{Models}

\paragraph{BART} is a denoising autoencoder and is pre-trained on a 160GB corpus of news, books, stories and webtext \cite {Liu2019RoBERTaAR}.  BART uses in-filling and sentence permutation noising functions.  Text infilling replaces a span of tokens with a single [MASK] token, while sentence permutation shuffles sub-sequences of sentences.  Encoder inputs are formed by infilling 30\% of the tokens in the input sequence and permuting all sentences.  The model is trained to a cross-entropy loss on the decoder’s ability to reconstruct the uncorrupted input.

We use the BART-Large model which consists of 12 layers, 16 attention heads, and a hidden dimension of 1024, yielding a 406MM parameter model.  The model uses beam search in generation with a beam width of 5 and a length penalty.

\paragraph{PEGASUS} is gap sentence generation model, in which an entire sentence is masked and the model aims to reconstruct the sentences from the surrounding context. It is pre-trained on the 750GB C4 and 3.8TB HugeNews corpora.  PEGASUS uses gap sentence masking as its noising function.  Entire sentences identified as important via heuristics are replaced with a gap-sentence-specific [MASK] token.  Encoder inputs are formed by masking gap sentences with ratios ranging from 15\% up to 75\%.  The model is trained to a cross-entropy loss on the decoder’s ability to reconstruct the masked gap sentences.

We use the PEGASUS-Large model which consists of 16 layers, 16 attention heads and a hidden dimension of 1024, yielding a 568MM parameter model.  
The model uses beam search for the summary generation with a beam width of 8 and a length penalty.

\paragraph{T5} is a text-to-text transfer learning model and is pre-trained on the 750GB C4 corpus using a noising function similar to infilling.  However, instead of replacing spans of tokens with a single [MASK] token, each span is replaced with a sentinel token which is unique to the sequence.  Encoder inputs are formed by replacing 15\% of the original tokens with sentinel tokens.  The model is trained to a cross-entropy loss in the decoder’s ability to reconstruct individual sentinel tokens.

We use the T5-Base model which consists of 12 layers, 12 attention heads and a hidden dimension of 768, yielding a 220MM parameter model.  
The model uses beam search for the summary generation with a beam width of 4 and a length penalty.

\input{rouge2.tex}

\section{Experimental Setup}

The three pre-trained models are fine-tuned on each dataset described above for three epochs with per-device batch sizes of 8 using default learning rates and an Adam optimizer using three Quadro-RTX 8000 GPUs.
During fine-tuning, models are optimized to a maximum likelihood objective for auto-regressive greedily decoded text for human written summaries. During testing, fine-tuned models decode summaries of the input text on a held-out test set using beam search. Each model used in this work truncates the summary at the specified target length. Each summarizer uses a different tokenizer, resulting in target lengths varying by model across each dataset. The width of the beam, length penalties, and the target summary lengths are hyper-parameters of the model.

\section{Full Results}

Table~\ref{tab:cross_domain_rouge} shows the detailed ROUGE-2 F1 scores for in-domain, cross-domain and multi-domain performance. 
The first six rows and columns in each panel make it easy to eyeball ROUGE-2 F1 scores for the in-domain and cross-domain performance of the same summarizer. 
The diagonal shows in-domain scores; off-diagonal entries are scores for cross-domain performance.  Without exception, the in-domain scores on the diagonal are markedly higher than the cross-domain scores. Fine-tuning with the mixed-domain training set results in a summarizer that has the best performance on the mixed-domain test set for all three models. The mixed-domain summarizer also achieves good scores for each domain, second best to the in-domain setting. 

The difference in performance for models fine-tuned on mixed domain and in-domain is small (see Table 3) to the point of being negligible. Remember however that the mix-domain fine-tuning set is much smaller than the in-domain fine-tuning sets. This finding highlights an inefficiency in creating the research datasets: they are much bigger than what appears to be necessary for practical good performance. The size entails a high price in time and computation for fine-tuning and inference. Ideally, the appropriate size of both fine-tuning and test set should be thoughtfully determined to optimize system performance and power to differentiate levels of system performance.
\section{Examples}
Table \ref{tab:examples} gives examples of human-generated summaries rated as inferior in information context along with the automatic summary, judged as superior by one or more human evaluators. 

\begin{table*}[th]
\small
\centering
\begin{tabular}{lrlr} \toprule
\multicolumn{2}{c}{Human Summary} & \multicolumn{2}{c}{Automatic Summary}\\ \cmidrule(lr){1-2} \cmidrule(lr){3-4}
text & rel-q & text & info \\ \cmidrule(lr){1-2} \cmidrule(lr){3-4}
\multirow{3}{0.4\linewidth}{A logic question about "Cheryl's birthday" goes viral . The clues give just enough information to eliminate most possibilities . It spread after a Singapore television host posted it to Facebook .} & 3.3 & \multirow{7}{0.4\linewidth}{Cheryl's birthday challenge was meant to test the better high-school students competing in the Singapore and Asian Schools Math Olympiad. Some are even saying it's the math equivalent of the "What color is the dress" debate. See our video below for the answer from Georgia Tech Mathematician Matt Baker.} & 4.7 \\ \\ \\ \\ \\ \\ \\ \midrule

\multirow{2}{0.4\linewidth}{The ramp agent fell asleep in the plane's cargo hold . He can no longer work on Alaska Airlines flights .} & 4.0 & \multirow{7}{0.4\linewidth}{911 call: "I'm inside a plane and I feel like it's up moving in the air" The caller was a ramp agent who fell asleep in the plane's cargo hold. The pilot radioed air traffic control and said he would make an emergency landing. All ramp employees have security badges, and undergo full criminal background checks before being hired.} & 5.0 \\ \\ \\ \\ \\ \\ \\ \midrule

\multirow{1}{0.4\linewidth}{my mom's face dislikes popsicles being thrown at it.} & 4.3 & \multirow{2}{0.4\linewidth}{tossed my mom a popsicle, hit her in the face, now she has a black eye.} & 5.0 \\ \\ \midrule

\multirow{2}{0.4\linewidth}{exploited my inability to get sunburns, turned out i was ``allergic'' to the sun and was destroying my body.} & 4.0 & \multirow{7}{0.4\linewidth}{Lupus is not the deadly kind, and it's only present on parts of my skin. However, i'm basically allergic to the sun. For the past 2 years i've essentially been making my body destroy \& repair itself repeatedly. For the past 2 years i've essentially been making my body destroy \& repair itself repeatedly. } & 4.7\\ \\ \\ \\ \\ \\ \\ \midrule

\multirow{1}{0.4\linewidth}{plugged in a customers portable dvd player the wrong way and f***ed it.} & 4.3 & \multirow{5}{0.4\linewidth}{A customer came in for a car adapter for her portable dvd player. She needed a 12v to 9v adapter. After plugging it in, a burning electric smell appeared. It was fried. Her boss had to give one of} & 4.0\\ \\ \\ \\ \\ \bottomrule
\end{tabular}
\caption{Example of human summaries rated as inferior in information context to some automatic summary, along with an example automatic summary with high information context score. rel-q is the score for the human summary on a scale from 1 (none) to 5 (many) denoting the extent to which automatic summaries have better content. info is the score for the automatic summary from 1 (no good info) to 5 (a lot of good info) denoting whether it provides good information. rel-q and info are not comparable.}
\end{table*}

\label{tab:examples}

\end{document}

%% file: dataset_statistics.tex
\begin{table*}[t]
\centering
\small
\begin{tabular}{llrrlr}
\toprule
Dataset & Domain & \# docs & doc len & summary src & sum len \\ \midrule
arXiv & scientific papers & 215k & 4938 & paper abstract & 220\\
Billsum & U.S. Congressional bills & 23k & 1382 & Congressional Research Service & 197\\
& California state legislative bills & & 1684 & state Legislative Counsel & \\
CNN/DailyMail & news & 300k & 781 & article bullet highlights & 56\\
GovReport & U.S. Govt reports & 19k & 9017 & experts & 542\\
PubMed & biomedical papers & 133k & 3016 & paper abstract & 203\\
TIFU & Reddit & 120k & 432 & post TL;DR & 23\\ 
Mixed-domain & All & 105k & \\ \bottomrule
\end{tabular}
\caption{Dataset statistics. Average lengths are in words.}
\label{tab:datasets}
\end{table*}

%% file: avg_auto_metrics.tex
\begin{table}[t]
\small
\centering
\setlength{\tabcolsep}{4pt}
\begin{tabular}{ll*3r} \toprule
& & BART & PEGASUS & T5\\  \midrule
\multirow{3}{1.8cm}{in-domain test} & ROUGE2 & {\bf 17.3} & 15.9 & 14.3\\
& BLEU & {\bf 12.9} & {\bf 12.9} & 11.8\\
& BERTscore & {\bf 89.7} & 89.0 & 88.6\\ \midrule
\multirow{3}{1.8cm}{cross-domain test} & ROUGE2 & {\bf 7.5} & 6.5 & 6.4\\
& BLEU & 2.7 & {\bf 2.8} & {\bf 2.8}\\
& BERTscore & {\bf 86.6} & 85.2 & 85.6\\ 
\bottomrule
\end{tabular}
\caption{Average automatic scores for in-domain, cross-domain and mixed-domain evaluation.
These scores exclude the mixed domain summarizer. Columns are the pre-trained models used. The highest score in each row is boldfaced.}
\label{tab:averages}
\end{table}

\begin{table}[t]
\small
\centering
\setlength{\tabcolsep}{3pt}
\begin{tabular}{l*3r} \toprule
& in-domain & CNN-DM & mixed-domain\\  \midrule
ROUGE2 & {\bf 17.3} & 7.5 & 15.7\\
BLEU & {\bf 12.9} & 2.7 & 9.6 \\
BERTscore & {\bf 89.7} & 87.3 & 89.5\\ 
\bottomrule
\end{tabular}
\caption{Average automatic scores on all test datasets for BART trained on different datasets. Columns are the training datasets used. in-domain is the average of scores with six models evaluated on their respective test splits or the mixed-domain test data. CNN and mixed-domain are single models evaluated on each test set.}
\label{tab:averages_data}
\end{table}

%% file: detail_auto_metrics.tex
\begin{table*}[t]
\centering
\small
\begin{tabular}{lllrrrrrrr} \toprule
& & & \multicolumn{7}{c}{Training Dataset}\\ \cmidrule(lr){4-10}
& & & arXiv & BillSum & CNN & Gov & PubMed & TIFU & Mixed\\ \midrule
\multirow{6}{*}{BART} & \multirow{3}{*}{in-domain} & ROUGE2 & 15.9 & 29.7 & 15.5 & 15.9 & 18.2 & 8.6 & 18.1\\
& & BLEU & 11.6 & 18.1& 13.8 & 11.8 & 16.3 & 5.9 & 10.4 \\
& & BERTscore & 89.2 & 90.6 & 90.1 & 88.9 & 88.9 & 90.5 & 89.9 \\ \cmidrule(lr){2-10}
& \multirow{3}{*}{Avg cross-domain $\Delta$} & ROUGE2 & -6.2 & -22.6 & -9.4 & -6.4 & -8.2 & -3.9 & -2.4\\
& & BLEU & -6.9 & -15.8 & -13.3 & -5.9 & -11.6 & -5.5 & -0.8\\
& & BERTscore & -1.9 & -3.7 & -3.2 & -2.5 & -1.3 & -5.4 & -0.4\\ \midrule
\multirow{6}{*}{T5} & \multirow{3}{*}{in-domain} & ROUGE2 & 12.2 & 30.2 & 13.7 & 7.3 & 16.1 & 6.2 & 16.7\\
& & BLEU & 8.2 & 25.5 & 12.3 & 5.4 & 15.3 & 3.8 & 11.0 \\
& & BERTscore & 87.3 & 90.3 & 90.0 & 86.5 & 87.7 & 89.8 & 88.8 \\ \cmidrule(lr){2-10}
& \multirow{3}{*}{Avg cross-domain $\Delta$} & ROUGE2 & -4.7 & -22.0 & -8.1 & -0.7 & -7.6 & -2.0 & -2.9\\
& & BLEU & -3.3 & -22.0 & -11.9 & -1.4 & -9.9 & -3.3 & -1.0\\
& & BERTscore & -2.6 & -3.2 & -3.4 & -1.1 & -1.8 & -5.3 & -0.5\\ \midrule
\multirow{6}{*}{PEGASUS} & \multirow{3}{*}{in-domain} & ROUGE2 & 13.6 & 30.7 & 14.4 & 11.0 & 18.2 & 7.7 & 16.6\\
& & BLEU & 9.8 & 24.3 & 12.0 & 8.5 & 17.7 & 4.8 & 11.0 \\
& & BERTscore & 87.9 & 90.3 & 89.8 & 87.6 & 88.3 & 90.1 & 88.9 \\ \cmidrule(lr){2-10}
& \multirow{3}{*}{Avg cross-domain $\Delta$} & ROUGE2 & -7.1 & -23.5 & -8.0 & -2.4 & -11.0 & -2.6 & -2.2\\
& & BLEU & -5.5 & -20.4 & -11.3 & -3.3 & -13.2 & -4.2 & -1.4\\
& & BERTscore & -3.7 & -4.9 & -2.9 & -0.8 & -3.5 & -6.0 & -0.4\\ 
\bottomrule
\end{tabular}
\caption{Scores for in-domain testing and the average degradation in the score w.r.t. in-domain score for out-of-domain testing. Columns represent models finetuned on each of the domains.}
\label{tab:cross_domain_agg}
\end{table*}

%% file: round1_human.tex
\begin{table}[t]
\centering
\small
\begin{tabular}{lrrr}
\toprule
model &  BART &  Pegasus &  T5 \\
\midrule
readability   & {\bf 3.97} & 3.70 & 3.46 \\
recall        & {\bf 3.72} & 3.42 & 3.07 \\
precision     & {\bf 1.48} & 1.89 & 2.66 \\
hallucination & {\bf 4.84} & 4.83 & 4.75 \\
orthography   & {\bf 0.37} & 0.29 & 0.27 \\
repetition    & {\bf 0.01} & 0.19 & 0.44 \\
\bottomrule
\end{tabular}
\caption{Human evaluation comparing the three models fine-tuned on mixed-domain data. A lower score is better for precision and repetition. A higher score is better for other dimensions.}
\label{tab:round1}
\end{table}

%% file: round2_human.tex
\begin{table}[t]
\centering
\small
\begin{tabular}{lrrr}
\toprule
model &  in-domain &  CNN-DM &  mixed \\
\midrule
readability   & 3.77 & {\bf 4.13} &  4.06 \\
recall        & 3.57 & 2.27 & {\bf 3.76} \\
precision     & 1.72 & 2.53 & {\bf 1.45} \\
hallucination & 4.86 & {\bf 4.89} & 4.85 \\
orthography   & 0.26 & {\bf 0.37} & 0.31 \\
repetition    & {\bf 0.01} & 0.02 & {\bf 0.01} \\
\bottomrule
\end{tabular}
\caption{Human evaluation comparing BART fine-tuned in-domain, CNN-DM and the mixed-domain datasets. A lower score is better for precision and repetition. A higher score is better for other dimensions.}
\label{tab:Round2}
\end{table}

%% file: rouge2.tex
\begin{table*}[t]
\centering
\small
\begin{tabular}{llrrrrrrr} \toprule
& & \multicolumn{7}{c}{Training Dataset}\\ \cmidrule(lr){3-9}
& & arXiv & BillSum & CNN & Gov & PubMed & TIFU & Mixed\\ \midrule
\multirow{7}{*}{BART} & arXiv & {\em 15.9} & 7.0 & 5.3 & 10.7 & 13.8 & 4.0 & 14.9\\
& BillSum & 14.2 & {\em 29.7} & 11.4 & 14.6 & 14.6 & 5.6 & 29.7\\
& CNN     & 8.0 & 8.2 & {\em 15.5} & 6.4 & 8.9 & 7.3 & 13.7\\
& Gov     & 8.3 & 7.9 & 3.0 & {\em 15.9} & 8.0 & 1.8 & 11.8\\
& PubMed  & 14.6 & 6.6 & 6.5 & 13.0 & {\em 18.2} & 3.9 & 15.9\\
& TIFU    & 1.9 & 1.4 & 3.0 & 1.8 & 2.5 & {\em 8.6} & 8.1\\
& Mixed    & 11.2 & 11.4 & 7.7 & 10.7 & 12.2 & 5.7 & {\em \underline{18.1}}\\ \midrule

\multirow{7}{*}{T5} & arXiv & {\em 12.2} & 8.0 & 4.8 & 6.7 & 9.9 & 2.7 & 11.1\\
& BillSum & 12.2 & {\em 30.2} & 8.6 & 10.8 & 14.6 & 4.1 & 30.1\\
& CNN & 6.5 & 10.9 & {\em 13.7} & 4.7 & 8.3 & 8.1 & 13.5\\
& GovReport & 6.2 & 6.1 & 3.1 & {\em 7.3} & 6.3 & 1.6 & 8.9\\
& PubMed & 10.0 & 9.8 & 6.8 & 8.7 & {\em 16.1} & 3.7 & 14.5\\
& TIFU & 0.8 & 2.1 & 2.7 & 0.9 & 1.4 & {\em 6.2} & 5.0\\
& Mixed & 9.2 & 12.6 & 7.4 & 7.7 & 10.5 & 5.1 & {\em \underline{16.7}}\\ \midrule

\multirow{7}{*}{PEGASUS} & arXiv & {\em 13.6} & 6.4 & 6.2 & 8.6 & 11.9 & 2.6 & 12.5 \\
& BillSum & 8.1 & {\em 30.7} & 10.8 & 12.2 & 8.1 & 6.4 & 30.4\\
& CNN & 3.7 & 8.0 & {\em 14.4} & 7.1 & 4.8 & 9.5 & 12.6\\
& GovReport & 4.7 & 5.5 & 3.2 & {\em 11.0} & 6.3 & 2.5 & 9.0\\
& PubMed & 13.5 & 9.0 & 7.7 & 12.5 & {\em 18.2} & 4.0 & 15.4\\
& TIFU & 1.4 & 2.3 & 2.4 & 2.0 & 2.1 & {\em 7.7} & 6.4\\
& Mixed & 7.7 & 11.9 & 7.9 & 9.5 & 10.1 & 5.7 & {\em \underline{16.6}}\\ \bottomrule
\end{tabular}
\caption{ROUGE-2 F1 Scores. Columns are training domains and rows are test domains.}
\label{tab:cross_domain_rouge}
\end{table*}

%% file: acl2023.bbl
\begin{thebibliography}{18}
\expandafter\ifx\csname natexlab\endcsname\relax\def\natexlab#1{#1}\fi

\bibitem[{Bar-Haim et~al.(2020)Bar-Haim, Kantor, Eden, Friedman, Lahav, and
  Slonim}]{bar-haim-etal-2020-quantitative}
Roy Bar-Haim, Yoav Kantor, Lilach Eden, Roni Friedman, Dan Lahav, and Noam
  Slonim. 2020.
\newblock \href {https://doi.org/10.18653/v1/2020.emnlp-main.3} {Quantitative
  argument summarization and beyond: Cross-domain key point analysis}.
\newblock In \emph{Proceedings of the 2020 Conference on Empirical Methods in
  Natural Language Processing (EMNLP)}, pages 39--49, Online. Association for
  Computational Linguistics.

\bibitem[{Cohan et~al.(2018)Cohan, Dernoncourt, Kim, Bui, Kim, Chang, and
  Goharian}]{Cohan2018ADA}
Arman Cohan, Franck Dernoncourt, Doo~Soon Kim, Trung Bui, Seokhwan Kim,
  W.~Chang, and Nazli Goharian. 2018.
\newblock A discourse-aware attention model for abstractive summarization of
  long documents.
\newblock In \emph{NAACL}.

\bibitem[{Hermann et~al.(2015)Hermann, Kocisk{\'y}, Grefenstette, Espeholt,
  Kay, Suleyman, and Blunsom}]{Hermann2015TeachingMT}
Karl~Moritz Hermann, Tom{\'a}s Kocisk{\'y}, Edward Grefenstette, Lasse
  Espeholt, Will Kay, Mustafa Suleyman, and Phil Blunsom. 2015.
\newblock Teaching machines to read and comprehend.
\newblock In \emph{Advances in Neural Information Processing Systems}.

\bibitem[{Hua and Wang(2017)}]{hua-wang-2017-pilot}
Xinyu Hua and Lu~Wang. 2017.
\newblock \href {https://doi.org/10.18653/v1/W17-4513} {A pilot study of domain
  adaptation effect for neural abstractive summarization}.
\newblock In \emph{Proceedings of the Workshop on New Frontiers in
  Summarization}, pages 100--106, Copenhagen, Denmark. Association for
  Computational Linguistics.

\bibitem[{Huang et~al.(2021)Huang, Cao, Parulian, Ji, and
  Wang}]{Huang2021EfficientAF}
Luyang Huang, Shuyang Cao, Nikolaus~Nova Parulian, Heng Ji, and Lu~Wang. 2021.
\newblock Efficient attentions for long document summarization.
\newblock In \emph{NAACL}.

\bibitem[{Kim et~al.(2019)Kim, Kim, and Kim}]{kim-etal-2019-abstractive}
Byeongchang Kim, Hyunwoo Kim, and Gunhee Kim. 2019.
\newblock \href {https://doi.org/10.18653/v1/N19-1260} {Abstractive
  summarization of {R}eddit posts with multi-level memory networks}.
\newblock In \emph{Proceedings of the 2019 Conference of the North {A}merican
  Chapter of the Association for Computational Linguistics: Human Language
  Technologies, Volume 1 (Long and Short Papers)}, pages 2519--2531,
  Minneapolis, Minnesota. Association for Computational Linguistics.

\bibitem[{Kornilova and Eidelman(2019)}]{kornilova-eidelman-2019-billsum}
Anastassia Kornilova and Vladimir Eidelman. 2019.
\newblock \href {https://doi.org/10.18653/v1/D19-5406} {{B}ill{S}um: A corpus
  for automatic summarization of {US} legislation}.
\newblock In \emph{Proceedings of the 2nd Workshop on New Frontiers in
  Summarization}, pages 48--56, Hong Kong, China. Association for Computational
  Linguistics.

\bibitem[{Lewis et~al.(2020)Lewis, Liu, Goyal, Ghazvininejad, Mohamed, Levy,
  Stoyanov, and Zettlemoyer}]{Lewis2020BARTDS}
Mike Lewis, Yinhan Liu, Naman Goyal, Marjan Ghazvininejad, Abdelrahman Mohamed,
  Omer Levy, Veselin Stoyanov, and Luke Zettlemoyer. 2020.
\newblock Bart: Denoising sequence-to-sequence pre-training for natural
  language generation, translation, and comprehension.
\newblock In \emph{ACL}.

\bibitem[{Lin(2004)}]{Lin2004ROUGEAP}
Chin-Yew Lin. 2004.
\newblock Rouge: A package for automatic evaluation of summaries.
\newblock In \emph{ACL 2004}.

\bibitem[{Liu et~al.(2019)Liu, Ott, Goyal, Du, Joshi, Chen, Levy, Lewis,
  Zettlemoyer, and Stoyanov}]{Liu2019RoBERTaAR}
Yinhan Liu, Myle Ott, Naman Goyal, Jingfei Du, Mandar Joshi, Danqi Chen, Omer
  Levy, Mike Lewis, Luke Zettlemoyer, and Veselin Stoyanov. 2019.
\newblock Roberta: A robustly optimized bert pretraining approach.
\newblock \emph{ArXiv}, abs/1907.11692.

\bibitem[{Luhn(1958)}]{luhn1958automatic}
Hans~Peter Luhn. 1958.
\newblock The automatic creation of literature abstracts.
\newblock \emph{IBM Journal of research and development}, 2(2):159--165.

\bibitem[{Post(2018)}]{Post2018}
Matt Post. 2018.
\newblock \href
  {https://www.amazon.science/publications/a-call-for-clarity-in-reporting-bleu-scores}
  {A call for clarity in reporting bleu scores}.
\newblock In \emph{WMT 2018}.

\bibitem[{Raffel et~al.(2020)Raffel, Shazeer, Roberts, Lee, Narang, Matena,
  Zhou, Li, and Liu}]{JMLR:v21:20-074}
Colin Raffel, Noam Shazeer, Adam Roberts, Katherine Lee, Sharan Narang, Michael
  Matena, Yanqi Zhou, Wei Li, and Peter~J. Liu. 2020.
\newblock \href {http://jmlr.org/papers/v21/20-074.html} {Exploring the limits
  of transfer learning with a unified text-to-text transformer}.
\newblock \emph{Journal of Machine Learning Research}, 21(140):1--67.

\bibitem[{Sandu et~al.(2010)Sandu, Carenini, Murray, and
  Ng}]{sandu-etal-2010-domain}
Oana Sandu, Giuseppe Carenini, Gabriel Murray, and Raymond Ng. 2010.
\newblock \href {https://aclanthology.org/W10-2603} {Domain adaptation to
  summarize human conversations}.
\newblock In \emph{Proceedings of the 2010 Workshop on Domain Adaptation for
  Natural Language Processing}, pages 16--22, Uppsala, Sweden. Association for
  Computational Linguistics.

\bibitem[{Wolf et~al.(2020)Wolf, Debut, Sanh, Chaumond, Delangue, Moi, Cistac,
  Rault, Louf, Funtowicz, and Brew}]{Wolf2020TransformersSN}
Thomas Wolf, Lysandre Debut, Victor Sanh, Julien Chaumond, Clement Delangue,
  Anthony Moi, Pierric Cistac, Tim Rault, R{\'e}mi Louf, Morgan Funtowicz, and
  Jamie Brew. 2020.
\newblock Transformers: State-of-the-art natural language processing.
\newblock In \emph{EMNLP}.

\bibitem[{Yu et~al.(2021)Yu, Liu, and Fung}]{yu-etal-2021-adaptsum}
Tiezheng Yu, Zihan Liu, and Pascale Fung. 2021.
\newblock \href {https://doi.org/10.18653/v1/2021.naacl-main.471}
  {{A}dapt{S}um: Towards low-resource domain adaptation for abstractive
  summarization}.
\newblock In \emph{Proceedings of the 2021 Conference of the North American
  Chapter of the Association for Computational Linguistics: Human Language
  Technologies}, pages 5892--5904, Online. Association for Computational
  Linguistics.

\bibitem[{Zhang et~al.(2020)Zhang, Zhao, Saleh, and Liu}]{Zhang2020PEGASUSPW}
Jingqing Zhang, Yao Zhao, Mohammad Saleh, and Peter~J. Liu. 2020.
\newblock Pegasus: Pre-training with extracted gap-sentences for abstractive
  summarization.
\newblock \emph{ArXiv}, abs/1912.08777.

\bibitem[{Zhang* et~al.(2020)Zhang*, Kishore*, Wu*, Weinberger, and
  Artzi}]{bert-score}
Tianyi Zhang*, Varsha Kishore*, Felix Wu*, Kilian~Q. Weinberger, and Yoav
  Artzi. 2020.
\newblock \href {https://openreview.net/forum?id=SkeHuCVFDr} {Bertscore:
  Evaluating text generation with bert}.
\newblock In \emph{International Conference on Learning Representations}.

\end{thebibliography}
